\documentclass[preprint,authoryear,12pt]{elsarticle}

\usepackage[english]{babel}
\usepackage{lineno}

\usepackage{amssymb}
\usepackage{amsthm}
\usepackage{amsmath}
\usepackage{mathtools}

\usepackage{booktabs}
\usepackage{adjustbox}
\usepackage[disable]{todonotes}

\usepackage[section]{placeins}
\usepackage{nameref}

\usepackage[none]{hyphenat}

\usepackage{nameref}

\newcommand*\mean[1]{\bar{#1}}

\usepackage{pgfplots}
\pgfplotsset{compat=1.14}

\usepackage{tikz}
\usepackage{xcolor}

\usepackage{threeparttable}

\usepackage{subcaption}

\journal{An Elsevier Journal}

\begin{document}

\begin{frontmatter}




\author[jacobs]{Atilla Ozgur}
\author[konya]{Sevim Seda Yamac}

\address[jacobs]{Department of Logistics and Mathematics, Jacobs University Bremen gGmbH ,Campus Ring 1, 28759 Bremen, Germany}
\address[konya]{Konya Food and Agriculture University, Faculty of Agriculture and Natural Sciences, Department of Plant Production and Technologies, Konya, Turkey}

\title{Modelling of daily reference evapotranspiration using deep neural network in different climates}

\begin{abstract}
Precise and reliable estimation of reference evapotranspiration ($ET_o$) is  an essential for the irrigation and water resources management. 
$ET_o$ is difficult to predict due to its complex processes.
This complexity can be solved using machine learning methods. 
This study investigates the performance of artificial neural network (ANN) and deep neural network (DNN) models for estimating daily $ET_o$. 
Previously proposed ANN and DNN methods have been realized, and their performances have been compared. 
Six input data including maximum air temperature ($T_{max}$), minimum air temperature ($T_{min}$), solar radiation ($R_n$), maximum relative humidity ($RH_{max}$), minimum relative humidity ($RH_{min}$) and wind speed ($U_2$) are used from 4 meteorological stations  (Adana, Aksaray, Isparta and Niğde) during 1999-2018 in Turkey.
The results have shown that our proposed DNN models achieves satisfactory accuracy for daily $ET_o$ estimation compared to previous ANN and DNN models.
The best performance has been observed with the proposed model of DNN with SeLU activation function (P-DNN-SeLU) in Aksaray with coefficient of determination ($R^2$) of 0.9934, root mean square error (RMSE) of 0.2073 and  mean absolute error (MAE) of 0.1590, respectively. 
Therefore, the P-DNN-SeLU model could be recommended for estimation of $ET_o$ in other climate zones of the world.
\end{abstract}


\end{frontmatter}


\section{Introduction}
\label{section-introduction}

Reference evapotranspiration ($ET_o$) is an essential hydrological component for the sustainable and efficient management of agricultural water resources and the optimum irrigation scheduling \citep{Huang2019Evaluation,Wu2019Daily,Yamac2020Estimation}. 
Many direct and indirect methods have been recommended to estimate $ET_o$.
The direct method of $ET_o$ estimation could be accomplished by water budget measurement (e.g. weighting lysimeters) or water vapor transfer methods (e.g. eddy covariance and Bowen ratio) \citep{Huang2019Evaluation}. 
Unfortunately, these methods are time consuming and costly.
Additionally, they have some spatial and temporal limitations \citep{Irmak2003Solar,Dinpashoh2006Study,Peng2017best}. 
As an alternative to the direct methods, mathematical models based on meteorological data provided by weather stations can be used to estimate $ET_o$ \citep{Tao2018Reference}.

The FAO-56 Penman-Monteith (FAO-56 PM) equation is recommended by the Food and Agriculture Organization (FAO) of the United Nations as a reference model for $ET_o$ estimate \citep{Allen1998Crop}.
The FAO-56 PM incorporates both the aerodynamics and thermodynamics aspects and has more accurate results compared to the other empirical methods \citep{Fan2018Evaluation, Wu2019Daily}. 
The FAO-56 PM model has been evaluated against various other methods under diverse areas, climates and time steps (daily, weekly and monthly). 
The results show that FAO-56 PM method has better performance than other empirical equations \citep{Pereira2015Crop,Lopez-Urrea2006Testing}. 
However, the FAO-56 PM requires numerous features for $ET_o$ estimation, including the geological variables such as elevation and latitude, and meteorological variables such as maximum and minimum temperature, maximum and minimum relative humidity, wind speed and net solar radiation \citep{Shiri2014Comparison,Feng2017Evaluation,Peng2017best}.
These requirements bring a major drawback to the application of the FAO-56 PM model. 
Due to the limited availability of meteorological data, mainly in developing countries, simplified empirical models with fewer requirements have been proposed  \citep{Valiantzas2013Simplified,Ahooghalandari2016Developing}, such as temperature based \citep{Hargreaves1985Reference}, mass transfer based \citep{Trabert2019Neue} and radiation based models \citep{Priestley1972Assessment}.
These simplified methods are more accurate for  monthly and weekly $ET_o$ estimation while they are less accurate on a daily $ET_o$ estimation \citep{Torres2011Forecasting}.

The estimation of $ET_o$ is considered as a complex and highly nonlinear dynamic process depending on quality of meteorological variables \citep{Wu2019Daily}.
However, it is usually difficult to develop accurate empirical models considering all these nonlinear and complicated processes, especially when some important input parameters are lacking.
Recently, machine learning methods have been widely used to estimate complex process of $ET_o$ estimation because these methods do not require knowledge of internal variables to solve non-linear and multi-variable functions \citep{Kisi2015Pan, Yamac2020Evaluation}. 
Thus, various machine learning methods have been suggested for estimation of $ET_o$.
Among these, (1) artificial neural networks (ANN) \citep{Antonopoulos2017Daily, Ferreira2019Estimation}, (2) support vector machines (SVM) \citep{Fan2018Evaluation, Ferreira2019Estimation}, (3) tree based assemble methods \citep{Kisi2016investigation, Fan2018Evaluation}, (4) boosting \citep{Fan2019Light} can be mentioned.

Because of promising results and enormous potential for image processing and data analysis, the Deep Neural Network (DNN) methods have become increasingly popular in recent years \citep{Kamilaris2018Deep}.
The DNN methods are actually improved versions of the ANN methods \citep{LeCun2015Deep}.
The ANN with single hidden layer is commonly called as multi-layer perceptrons (MLP) or feed forward neural networks, while the ANN with more than two hidden layers are called Deep Neural Networks.
DNNs are interchangeably called  as deep neural networks, deep learning methods, or deep neural nets.
The DNN methods have been applied to different domains, such as speech recognition \citep{Amodei2016Deep}, natural language processing \citep{Young2018Recent}, and game playing \citep{Guo2014Deep}.
Likewise, the use of DNN methods recently increased in the area of hydrological \citep{Wang2020Deep, Lee2020Stochastic, Bui2020Verification} and agricultural \citep{Golhani2018review,Grinblat2016Deep,Dyrmann2016Plant} studies.

The current study makes following contributions: 
(1) Previously proposed methods are categorized in Table~\ref{table-literature-neural-networks} and Table~\ref{table-datasets-reviewed-studies}.
(2) Previously proposed methods have been realized and their performances have been compared in current meteorological dataset (Table~\ref{table-experimental-results-previous}).
(3) Different from previously proposed DNN method \citep{Saggi2019Reference}, dropout layer has been used and its performance is measured.
(4) Moreover, two activation function which are called rectifier linear units (ReLU) and scaled exponential linear units (SeLU) has been used in DNN method and compared with the other methods.
(5) All methods (previous and new) are compared using 5-fold cross validation instead of using single train-test dataset split.  
In 5-fold cross validation, models were trained on dataset with 5 different splits.

\section{Related works}
\label{section-related-works}

In this section, previous studies were reviewed for estimation of daily $ET_o$ value. 
Table~\ref{table-literature-neural-networks} summarizes the previous ANN methods in the literature. 
Table~\ref{table-datasets-reviewed-studies} demonstrates the dataset information of the previous ANN methods.

Various ANN methods were proposed for estimation of $ET_o$ value (Table~\ref{table-literature-neural-networks}). 
However, only two DNN methods  were proposed in the literature \citep{Saggi2019Reference,Ferreira2020New}.
The first DNN method used only 3 hidden layers with Rectifier Linear Units (ReLU) \citep{Saggi2019Reference}.
The second DNN method employed convolutional neural networks on hourly data \citep{Ferreira2020New}.
Since the currently proposed DNN methods use daily data, they are not comparable to the second DNN method.

\begin{table}[!htb]
\caption{Previously proposed neural network models for $ET_o$ estimation}
\label{table-literature-neural-networks}

\begin{adjustbox}{max width=\textwidth,max totalheight=\textheight,keepaspectratio}

\begin{tabular}{p{0.03\textwidth}p{0.3\textwidth}p{0.3\textwidth}p{0.2\textwidth}p{0.3\textwidth}}

\toprule
& Study & Neural Network Architecture & Activation Functions  &  Software \\ 

\midrule

1 & \cite{Landeras2008Comparison} & (4-6)–(1-14)-1 & NI & Statistica \\

2 & \cite{Traore2010Artificial} & (3-5)-(1-20)-1 & Sigmoid & NeuroSolution  \\

3 & \cite{Huo2012Artificial} & 5-8-1, 3-4-5-1, 4-5-6-1, 3-4-5-1 & Sigmoidal logistic  & Matlab  \\

4 & \cite{Rahimikhoob2014Comparison} & 4-(1-10)-1 & Log–sigmoid & Weka \\

5 & \cite{Shiri2014Comparison} & (4-5)-(1-14)-1 & Sigmoid, linear & Matlab \\

6 & \cite{Gocic2015Soft} & 5-3-6-10-1 & Continuous log–sigmoid & Matlab \\

7 & \cite{Kisi2016investigation} & 4-(3-9)-1,2-(3-10)-1 & Sigmoid, linear & NI \\

8 & \cite{Yassin2016Artificial} & (4-9)-(2-20)-1 & Sigmoid, linear & Multiple Back-Propagation \\

9 & \cite{Feng2016Comparison} & (2,3,5)-6-1 & NI & Matlab \\

10 & \cite{Antonopoulos2017Daily} &   (2-4)-6-1, 2-4-1 & Sigmoid & NI \\

11 & \cite{Dou2018Evapotranspiration} & 4-11-1, 4-15-1 & Sigmoid, linear & Matlab  \\

12 & \cite{Saggi2019Reference} & 7-40-60-40-1 & ReLU, softmax  & $H_2O$ \\

 \bottomrule

\end{tabular}
\end{adjustbox}
\end{table}

As can be seen in Table~\ref{table-literature-neural-networks}, all of the architectures had standard input layer, hidden layer(s) and output layer.
For example, architecture of \cite{Landeras2008Comparison} was "(4-6)–(1-14)-1".
This means that they used 4 to 6 neurons in the input layer, 1 to 14 neurons in the single hidden layer and 1 neuron in the output layer.
They performed empirical experiments and reported their best results among the tried number of neurons.
Also, other studies performed similar empirical experiments \citep{Landeras2008Comparison,Traore2010Artificial,Rahimikhoob2014Comparison,Shiri2014Comparison,Kisi2016investigation,Yassin2016Artificial}.
As an another example, \cite{Gocic2015Soft} used architecture of "5-3-6-10-1".
This means that they used 5 neurons in the input layer, 3-6-10 neurons in the 3-hidden layers and 1 neuron in the output layer.
Since model of \cite{Gocic2015Soft} used more than 2 layers in the architecture, they could have chosen to call their method DNN but they did not.
This could be the fact that they did not use any other DNN improvements in their experiments.

Activation functions that are used in literature are given in the third column (Table~\ref{table-literature-neural-networks}).
All the previous studies used standard sigmoid and linear functions except for the previously applied DNN method \citep{Saggi2019Reference}.
In addition, last column shows the software that was used in studies.
Most of the studies (5/12) used Matlab neural network toolbox.
Unfortunately, some studies did not report the activation functions and software, making reproducibility of their studies harder if not impossible.

\begin{table}[!htb]
\centering
\caption{Dataset information of the previously proposed neural network methods for $ET_o$ estimation}
\label{table-datasets-reviewed-studies}
\begin{adjustbox}{width=\textwidth}

\begin{tabular}{p{0.03\textwidth}p{0.2\textwidth}p{0.15\textwidth}p{0.15\textwidth}p{0.5\textwidth}}

\toprule
 & Study      & Years & Frequency  &  Train Validation Test   \\ 

\midrule

1 & \cite{Landeras2008Comparison}  & 1999-2003 &  Daily & 1999-2001 Train (75\%) validation (25\%), 2002-2003 Test   \\

2 & \cite{Traore2010Artificial} & 1996-2006 & Daily & 1996-2003 Train, 2004-2005 Validation, 2006 Test  \\

3 & \cite{Huo2012Artificial}  & 1952-2001 &  Daily & 1952-1986 Train, 1987-2001 Test   \\

4 & \cite{Rahimikhoob2014Comparison}  & 1998-2007 & Monthly & 1998-2004 Train, 2005-2007 Test  \\ 

5 & \cite{Shiri2014Comparison} & 2000-2008 & Daily & 2000-2005 Train, 2006-2008 Test  \\

6 & \cite{Gocic2015Soft}  & 1980-2010 &  Monthly & 1980-1995 Train, 1996-2010 Test   \\

7 & \cite{Kisi2016investigation}  & 1994-2009 &  Daily & 1998-2001 Train, 2002-2005 Validation, 2003-2009 Test   \\

8 & \cite{Yassin2016Artificial} & 1980-2010 & Daily &  Train (65\%), Validation (35\%) in 13 stations, Test using separate 6 stations   \\

9 & \cite{Feng2016Comparison} & 1994-2013 & Daily & 65\% Train, 35\% Test   \\

10 & \cite{Antonopoulos2017Daily} & 2009–2013 & Daily & 1 year Train, other 4 years Test \\

11 & \cite{Dou2018Evapotranspiration}  & 2001-2009 but 6 years of data used & Daily & 4 years Train, 1 year Validation, 1 year Test  \\

12 & \cite{Saggi2019Reference} & 1978–1999 and 2007–2016 H, 1970–1999 and 2007–2016 P & Daily & 55\% Train, 30\% Validation, 15\% Test \\

\bottomrule
\end{tabular}
\end{adjustbox}

\end{table}

Table~\ref{table-datasets-reviewed-studies} gives information about the years, frequency and dataset split (train, validation and test)  in the literature. 
It was important to show the split of datasets (train, validation and test) because this split affects the machine learning performances.
For example, \cite{Traore2010Artificial} reported that they used the meteorological dataset between 2004 and 2005 years as cross validation to optimize ANN performance.
However, this usage was for validation dataset. Similar usage were done also by \cite{Landeras2008Comparison,Yassin2016Artificial}.

Among the reviewed studies, 10 of 12 studies used the daily data for estimation of $ET_o$, while the other 2 studies used the monthly data.
Interestingly, no study evaluated their approaches using true cross validation in their experiments.
According to the best knowledge of the authors, the present study was the first study that uses true cross validation in the literature of $ET_o$ estimation.

\section{Materials and methods}
\label{section-materials-and-methods}

In this study, two newly DNN models were proposed to estimate daily value of $ET_o$. 
These newly proposed models were called P-DNN-ReLU and P-DNN-SeLU that uses activation ReLU and SeLU functions.
Previous DNN and ANN models were reproduced and their results were also included in experiments.
The dropout layer was also tried on all DNN models, though its effect on performance was not good.

In addition to current (P-DNN-ReLU and P-DNN-SeLU) and previous DNN model \citep{Saggi2019Reference}, 11 previous ANN methods were also implemented \citep{Landeras2008Comparison,Traore2010Artificial,Huo2012Artificial,Rahimikhoob2014Comparison,Shiri2014Comparison,Gocic2015Soft,Kisi2016investigation,Yassin2016Artificial,Feng2016Comparison,Antonopoulos2017Daily,Dou2018Evapotranspiration}.
The ANN methods used neuron size between 1 to 30 in their hidden layers. 
Therefore, 30 different ANN models were trained for every station.
For the DNN methods, the dropout layer was also used.
Therefore, $(6 \times 6 \times 6) \times 3 = 648$ different DNN models were trained also for every station.
In total, 678 models were trained in the present study.
Finally, considering 4 stations and 5 cross validation, the experiments trained and tested $678 \times 5 \times 4 = 13560$ different models.
All of these experimental results are available as a supplementary data. 
The flowchart of the modeling procedure is presented in Figure~\ref{figure-experiments-flowchart}.

\subsection{Background knowledge}
\label{section-background}

\begin{figure}[!hbt]

 \includegraphics[width=0.9\textwidth]{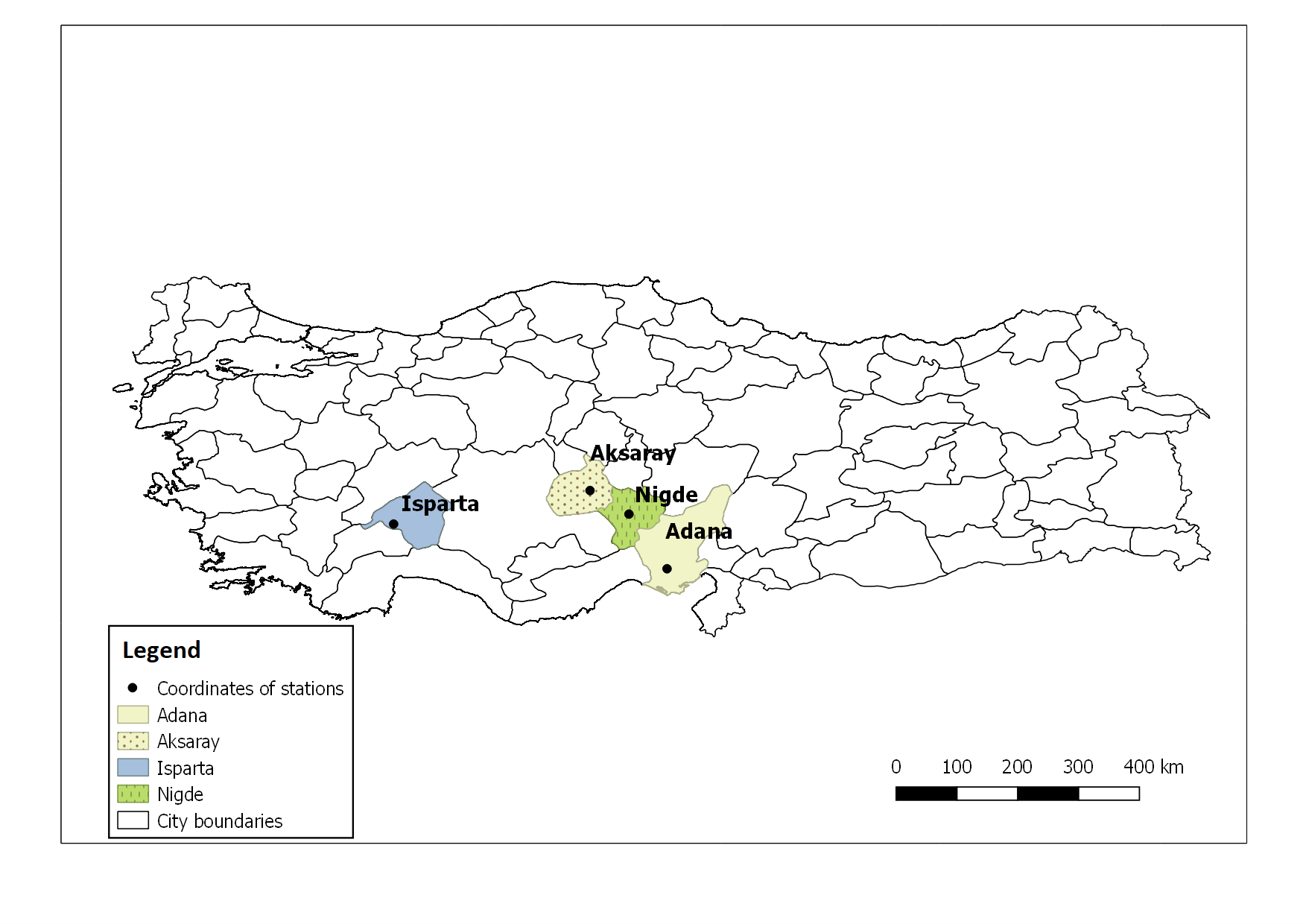}

\caption{Geographical locations of the 4 meteorological stations in Turkey.}
\protect{\label{figure-stations-turkey-map}}

\end{figure}

\subsubsection{Study area and dataset description}
\label{section-study-area-and-dataset-description}

The daily data from 4 meteorological stations in Turkey were obtained from Turkish State Meteorological Service for the period of 1999-2018.
Data features are maximum air temperature ($T_{max}$), minimum air temperature ($T_{min}$), solar radiation ($R_n$), maximum relative humidity ($RH_{max}$), minimum relative humidity ($RH_{min}$) and wind speed ($U_2$).
Table~\ref{table-dataset-summary} shows the statistical parameters of meteorological variables at Adana, Aksaray, Isparta and Niğde sites. 
The map of the study area and the location of the 4 meteorological stations are shown in Figure~\ref{figure-stations-turkey-map}.

According to the Köppen Geiger climate classification \citep{Kottek2006World}, the climate of Adana and Isparta sites have a warm temperature with a dry summer, while the climate of Aksaray and Niğde sites have a semi-arid with cold and snowy winters. 
In this way, the meteorological data collected from 4 different sites in Turkey was applied to two different climate types in this study. 
Table~\ref{table-dataset-summary} sums up the geographical and meteorological information of the 4 stations in Turkey.

\begin{table}[!htb]
\caption{Geographic information and statistical parameters of meteorological variables of the 4 stations (Adana, Aksaray, Isparta and Niğde) in Turkey during 1999-2018}
\label{table-dataset-summary}

\begin{adjustbox}{max width=\textwidth,max totalheight=\textheight,keepaspectratio}
\begin{threeparttable}[b]

\begin{tabular}{@{}lllllllllll@{}}
\toprule
Station & Station Code & Longitude & Latitude & Altitude & Variable & Min & Max & Mean & Std \tnote{1} & Cv \tnote{2} \\ 
\midrule

Adana & 1 & 35.34 & 37.00 & 23 & \multicolumn{6}{l}{}\\
\multicolumn{5}{l}{}  &  &     &     &      &     &   \\
\multicolumn{5}{l}{}  &  $T_{max}$  & 5.30 & 44.40 & 25.74 & 7.68 & 0.30  \\
\multicolumn{5}{l}{}  &  $T_{min}$  & -3.20 & 29.80 & 14.85 & 7.28 & 0.49  \\
\multicolumn{5}{l}{}  &  $R_s$  & 0.00 & 33.68 & 15.90 & 6.98 & 0.44  \\
\multicolumn{5}{l}{}  &  $R_{Hmax}$  & 27.00 & 100.00 & 86.22 & 11.38 & 0.13  \\
\multicolumn{5}{l}{}  &  $R_{Hmin}$  & 0.00 & 96.00 & 42.57 & 17.09 & 0.40  \\
\multicolumn{5}{l}{}  &  $U_2$  & 0.30 & 6.00 & 1.62 & 0.70 & 0.43  \\
\multicolumn{5}{l}{}  &  $ET_0$  & 0.51 & 12.73 & 4.45 & 2.20 & 0.50  \\
Aksaray & 2 & 34.00 & 38.37 & 970 & \multicolumn{6}{l}{}\\
\multicolumn{5}{l}{}  &  &     &     &      &     &   \\
\multicolumn{5}{l}{}  &  $T_{max}$  & -10.00 & 40.00 & 19.30 & 10.06 & 0.52  \\
\multicolumn{5}{l}{}  &  $T_{min}$  & -20.40 & 25.60 & 7.09 & 8.05 & 1.14  \\
\multicolumn{5}{l}{}  &  $R_s$  & 0.69 & 32.43 & 16.98 & 7.82 & 0.46  \\
\multicolumn{5}{l}{}  &  $R_{Hmax}$  & 20.00 & 100.00 & 71.40 & 16.46 & 0.23  \\
\multicolumn{5}{l}{}  &  $R_{Hmin}$  & 0.00 & 98.00 & 37.69 & 17.01 & 0.45  \\
\multicolumn{5}{l}{}  &  $U_2$  & 0.30 & 5.92 & 1.58 & 0.69 & 0.44  \\
\multicolumn{5}{l}{}  &  $ET_0$  & 0.34 & 10.61 & 4.24 & 2.55 & 0.60  \\
Isparta & 3 & 30.57 & 37.78 & 997 & \multicolumn{6}{l}{}\\
\multicolumn{5}{l}{}  &  &     &     &      &     &   \\
\multicolumn{5}{l}{}  &  $T_{max}$  & -6.60 & 42.30 & 19.24 & 9.49 & 0.49  \\
\multicolumn{5}{l}{}  &  $T_{min}$  & -16.00 & 23.30 & 6.53 & 7.19 & 1.10  \\
\multicolumn{5}{l}{}  &  $R_s$  & 0.00 & 32.52 & 15.30 & 7.99 & 0.52  \\
\multicolumn{5}{l}{}  &  $R_{Hmax}$  & 14.00 & 100.00 & 81.44 & 12.94 & 0.16  \\
\multicolumn{5}{l}{}  &  $R_{Hmin}$  & 0.00 & 99.00 & 40.93 & 16.87 & 0.41  \\
\multicolumn{5}{l}{}  &  $U_2$  & 0.00 & 5.78 & 1.32 & 0.70 & 0.53  \\
\multicolumn{5}{l}{}  &  $ET_0$  & 0.42 & 9.60 & 3.73 & 2.30 & 0.62  \\
Niğde & 4 & 34.68 & 37.96 & 1211 & \multicolumn{6}{l}{}\\
\multicolumn{5}{l}{}  &  &     &     &      &     &   \\
\multicolumn{5}{l}{}  &  $T_{max}$  & -10.30 & 38.50 & 18.51 & 9.90 & 0.53  \\
\multicolumn{5}{l}{}  &  $T_{min}$  & -19.80 & 23.00 & 6.02 & 7.89 & 1.31  \\
\multicolumn{5}{l}{}  &  $R_s$  & 0.68 & 35.10 & 18.75 & 8.38 & 0.45  \\
\multicolumn{5}{l}{}  &  $R_{Hmax}$  & 24.00 & 104.00 & 75.55 & 15.13 & 0.20  \\
\multicolumn{5}{l}{}  &  $R_{Hmin}$  & 2.00 & 96.00 & 37.44 & 17.49 & 0.47  \\
\multicolumn{5}{l}{}  &  $U_2$  & 0.38 & 7.95 & 1.83 & 0.70 & 0.38  \\
\multicolumn{5}{l}{}  &  $ET_0$  & 0.39 & 10.99 & 4.49 & 2.67 & 0.59  \\

\bottomrule
\end{tabular}
\begin{tablenotes}
\item [1] Std: Standard deviation
\item [2] Cv: Covariance of variance

\end{tablenotes}
\end{threeparttable}

\end{adjustbox}
\end{table}

\subsubsection{FAO Penman-Monteith equation}
\label{section-FAO Penman-Monteith-equation}
The FAO Penman-Monteith (FAO-56 PM) equation was proposed by \cite{Allen1998Crop}.
It is used to predict daily $ET_o\;(mm\;day^{-1})$ and provided the reference data for the training and testing in the current study.

\begin{equation}
    ET_o = \frac{0.408(R_n-G)+\gamma\frac{900}{T+273}U_2(e_s-e_a)}{\Delta+\gamma(1+0.34\;U_2)}
    \label{eq-fao-1}
\end{equation}

\noindent where $ET_o$ is the reference evapotranspiration $(mm\;day{^-1})$. 
$R_n$ is the net solar radiation $(MJ\;m^{-2}\;day^{-1})$, 
$G$ is the soil heat flux density $(MJ\;m^{-2}\;day^{-1})$, 
$T$ is the mean daily air temperature at 2m height $(^\circ{C})$, 
$\Delta$ is the slope of the saturated vapour pressure curve $(kPa\;^\circ{C}^{-1})$, 
$\gamma$ is the psychometric constant $(0.066\;kPa\;^\circ{C}^{-1})$, 
$e_s$ and $e_s$ are saturated and prevailing actual vapour pressure $(kPa)$, respectively, 
and $U_2$ is the mean daily wind speed $(m\;s^{-1})$ measured at 2 m height.

The saturation vapor pressure $e_s$ was estimated as:

\begin{equation}
    e_s=\frac{e^o(T_{max})+e^o(T_{min})}{2}
    \label{eq-fao-2}
\end{equation}

 where $e^o(T)$ is the saturation vapour pressure $(kPa)$ at the air temperature $(T)$, and $T_{max}$ and $T_{min}$ are maximum and minimum daily air temperature $(^\circ{C})$, respectively.
The saturation vapor pressure at the air temperature T was calculated as:
 
 \begin{equation}
    e^o(T)=0.6108\;e^{\left[\frac{17.27\;T}{T+237.3}\right]}
    \label{eq-fao-3}
\end{equation}

The actual vapor pressure $(e_a)$ was calculated as:

\begin{equation}
    e_a=\frac{e^o(T_{min})\frac{RH_{max}}{100}+e^o(T_{max})\frac{RH_{min}}{100}}{2}
    \label{eq-fao-vapor}
\end{equation}

where $RH_{max}$ is the maximum daily relative humidity, $(RH_{min})$ is the minimum daily relative humidity, $e^o(T_{min})$ is the saturation vapor pressure (kPa) at the minimum daily air temperature and $e^o(T_{max})$ is the saturation vapor pressure (kPa) at the maximum daily air temperature, respectively.

\subsubsection{Artificial neural networks and deep neural networks}
\label{section-neural-networks}

\begin{figure}[hbt!]

 \includegraphics[width=0.9\textwidth]{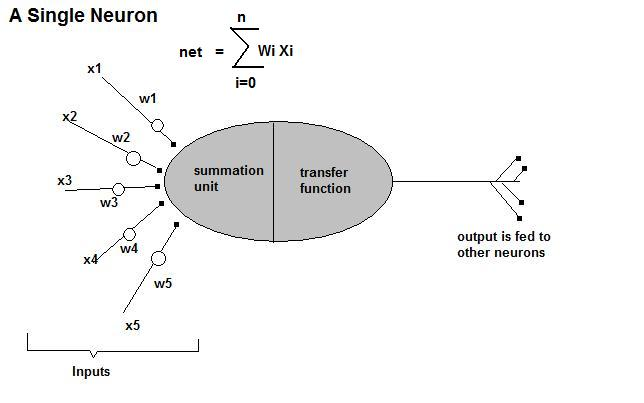}

\caption{Visualization of the artificial neuron.}
\protect{\label{figure-artificial-neuron}}

\end{figure}

Artificial neural networks (ANN) are powerful machine learning methods that take their roots from biological neurons. 
The ANN uses artificial neurons modeled from biological neurons as fundamental building block. 
An artificial neuron has 3 characteristics.
(1) inputs, (2) summation unit and (3) transfer (activation) function (Figure~\ref{figure-artificial-neuron}).

In the estimation of $ET_o$, inputs  are features used in the FAO-56 PM equation.
Inputs are multiplied with weights and added together in summation unit.
The summation value is sent to activation function and the output of this activation function is the output of the neuron.
Diverse transfer functions are proposed in the neural network literature \citep{Nwankpa2018Activation}.
The most common activation functions are sigmoid, Gaussian and linear.
Different transfer function are used for different purposes. 
For example, sigmoid function is used for binary classification whereas linear function is used for regression.

The ANN method is the umbrella term for machine learning methods that use neurons as building blocks.
Nonetheless, most studies use terms of Neural Networks, Artificial Neural Networks (ANN), Multi Layer Perceptrons (MLP) and Feed Forward Neural Networks interchangeably.
In the current study, artificial neural network term is preferred since this term is commonly used in related literature. 

\begin{figure}[hbt!]

 \includegraphics[width=0.9\textwidth]{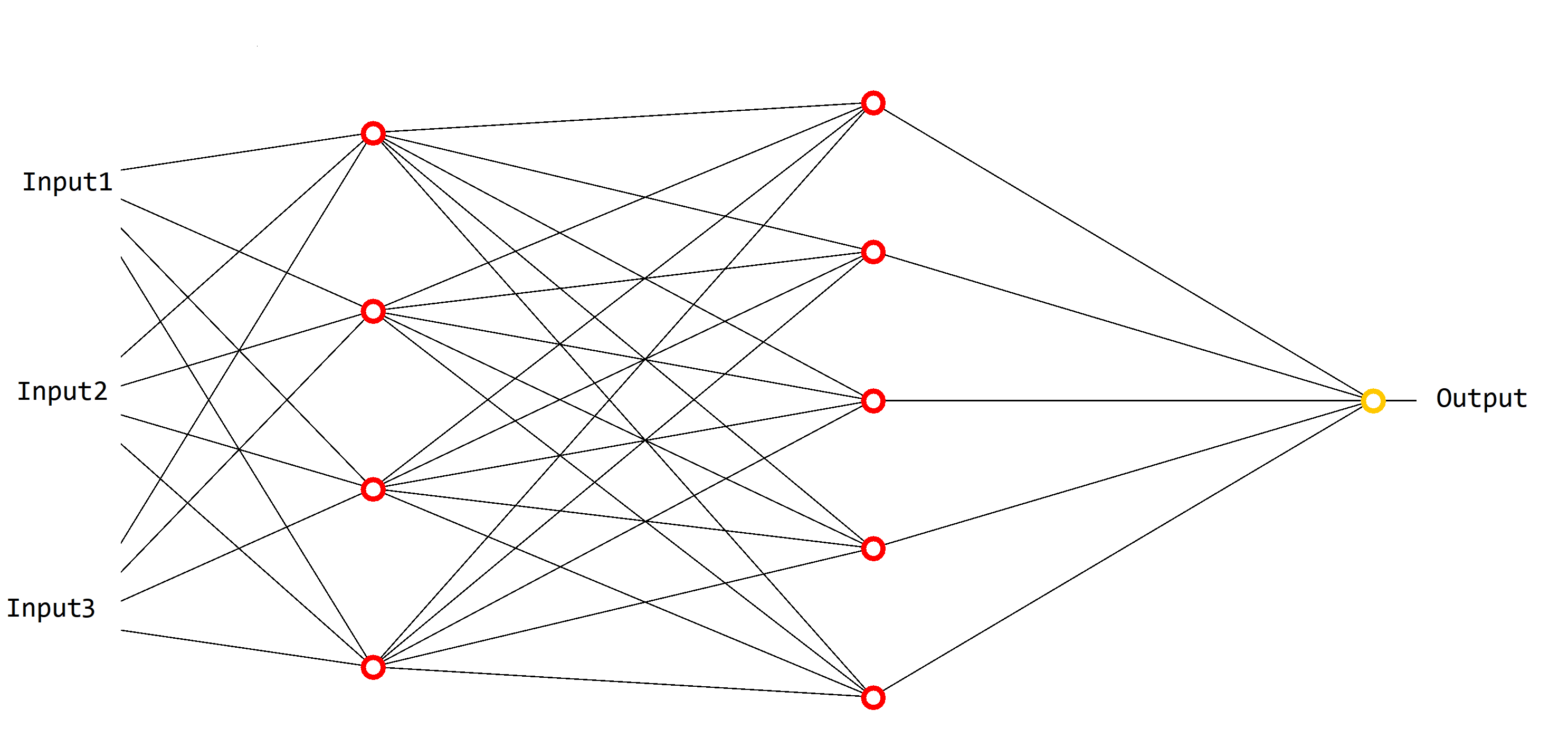}

\caption{Visualization of Neural Network (3-4-5-1) architecture (Note: Figure is created with Weka \citep{Hall2009WEKA}).}
\protect{\label{figure-neural-network-example}}

\end{figure}

In the ANN architecture, multiple neurons are used in multiple layers.
Most of the time, ANN architecture is identified with layer neuron counts such as $3-5-1$.
Here, it has 3 layers (one input, one hidden and one output layer) and there are 3 neurons in input layer, 5 neurons in one hidden layer and 1 neuron in output layer.
According to number of features used, input layer have corresponding number of neurons. 
An example for an ANN architecture is depicted in Figure~\ref{figure-neural-network-example}.

\subsubsection{Deep neural networks}
\label{section-deep-learning-methods}

Deep neural networks (DNN) are advanced versions of ANN methods \citep{Nielsen2015Neural,LeCun2015Deep}.
However, the differences between the classical ANN and DNN methods are not clearly defined in the literature, but following improvements are mostly related to DNN studies.

\begin{enumerate}
	\item Using more than 2 hidden layers, so called deep layers.
    \item Different neuron types, such as convolutional, pooling and dropout.
	\item Introduction of new activation functions, such as rectified linear units (ReLU), softmax and scaled exponential linear units (SeLU) functions and many others.
	\item Different training methods for back propagation that are more suited to using parallelization using multiple GPUs and CPUs.
	\item Different initialization methods for neuron weights.

\end{enumerate}

\subsubsection{Cross validation}
\label{section-train-validation-test-cross-validation}

\begin{figure}[!htb]
    \begin{center}
\fbox{
\includegraphics[width=8cm,height=4cm]{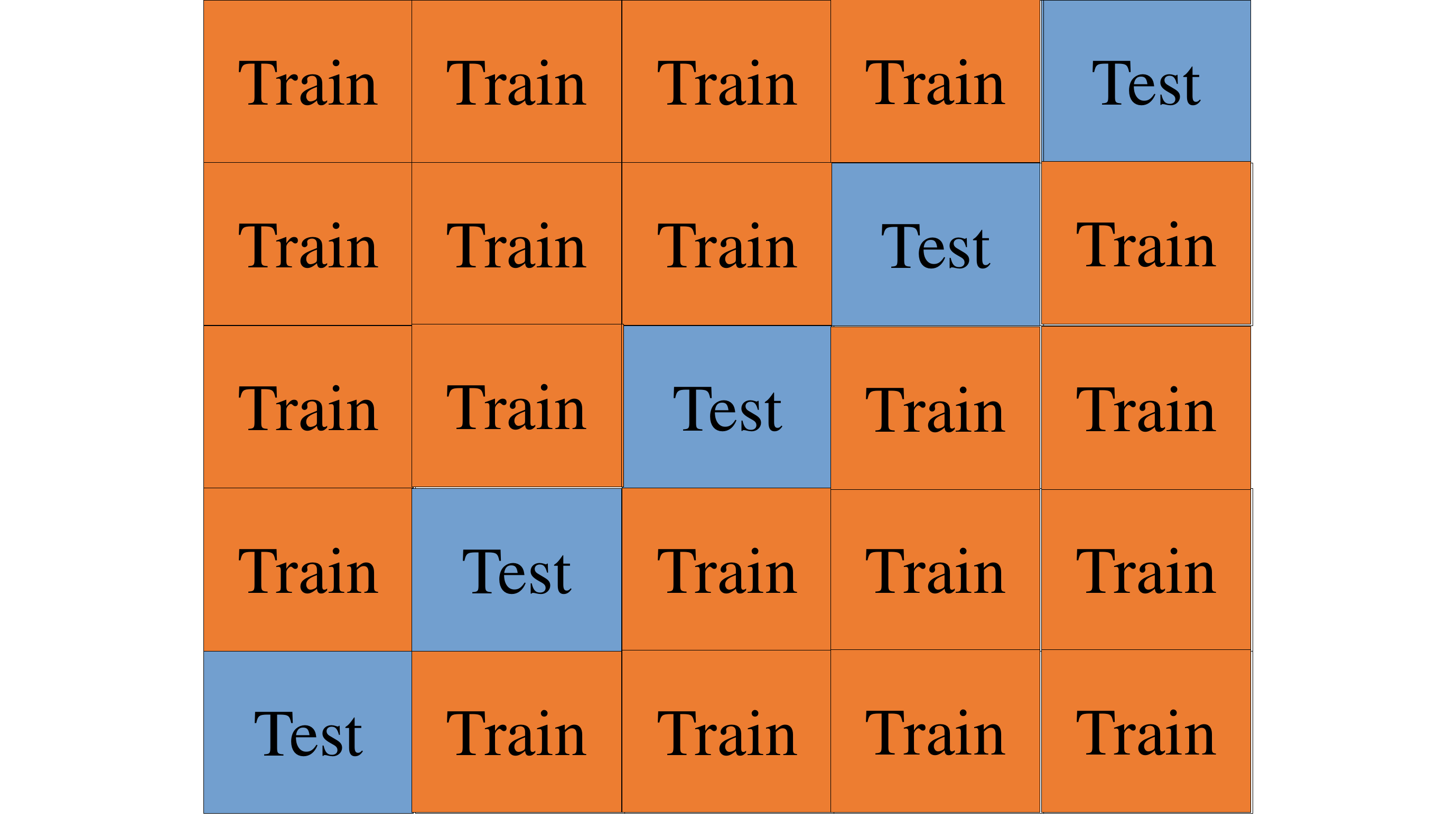}
}
    \end{center}
\caption{The visualization of 5-fold cross validation.}
\label{figure-cross-validation-5fold}
\end{figure}

\begin{figure}[hbt!]

 \includegraphics[width=1.0\textwidth]{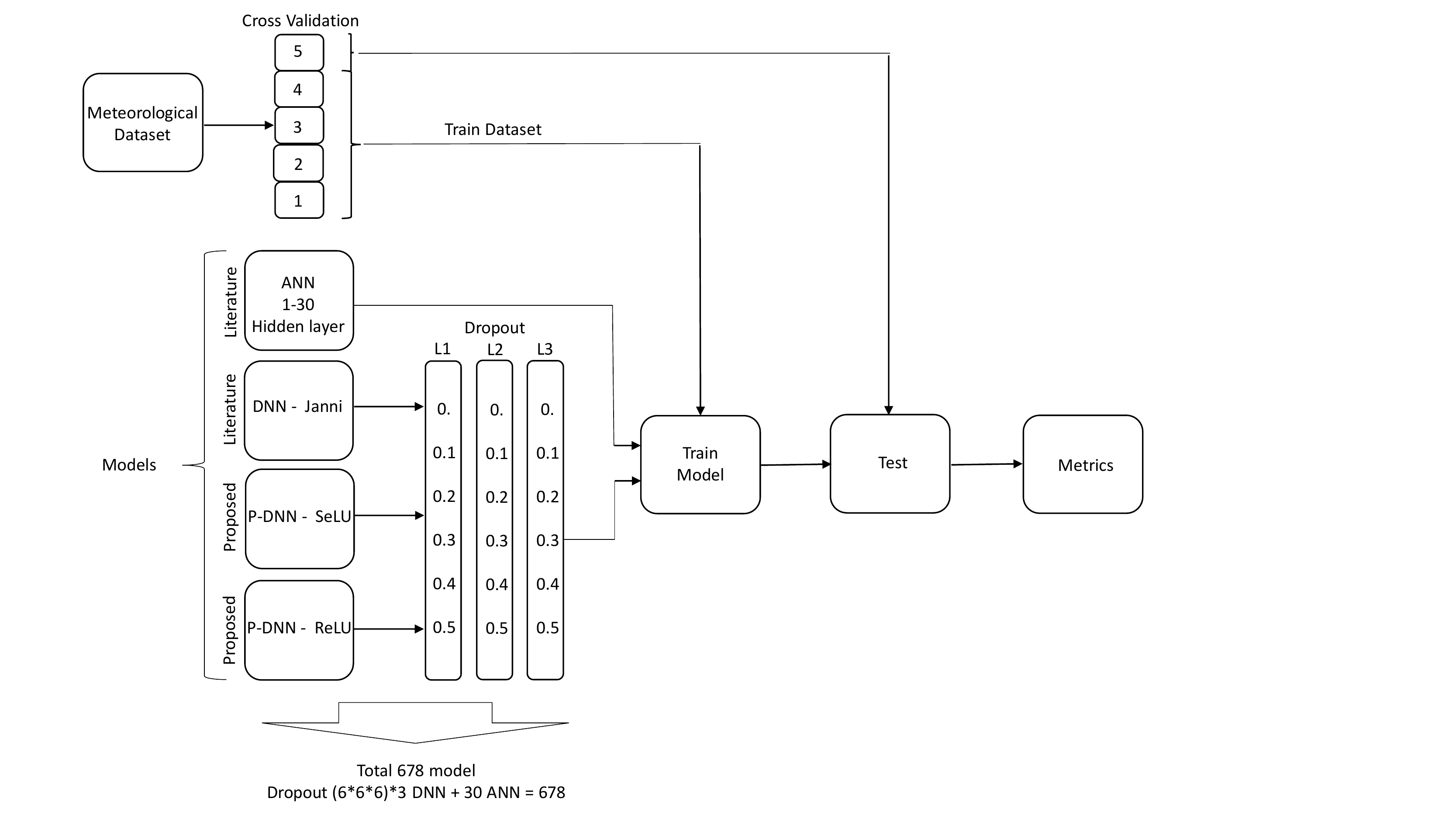}

\caption{The flowchart of the experimental procedures}
\label{figure-experiments-flowchart}

\end{figure}

The k-fold cross validation (\mbox{$k=5$}) method was used in this study.
The data were divided into k parts and algorithms were trained using \mbox{$k-1$} parts. 
After that the trained model was tested on remaining 1 part.
This procedure was repeated k times.
As an example, in the 5 fold cross validation, dataset was divided into 5 parts.
Using 4 parts, ANN and DNN models were trained, then tested on remaining 1 part.
This procedure was repeated 5 times (Figure~\ref{figure-cross-validation-5fold}).
Since experiments were repeated 5 times and performance metrics averaged, obtained performance metrics are more reliable than using only one train-test split.

\subsubsection{Performance evaluation of model parameters}
\label{Performance-evaluation-of-model-parameters}

The performance of the models were evaluated using the root mean square error (RMSE), mean absolute error (MAE), coefficient of determination $(R^2)$ in the training and testing subsets.

The daily $ET_o$ values, generated by the models $(S_i)$, were transformed into daily errors, comparing them with observed values $(O_i)$. $\mean{O}$ is the mean value of the observed values and $\mean{S}$ is the mean value of the observed values.

\begin{equation}
    RMSE=\sqrt{\frac{\sum_{i=1}^n(S_i-O_i)^2}{n}}
\end{equation}

\begin{equation}
    MAE=\left|{\frac{\sum_{i=1}^n(S_i-O_i)}{n}}\right|
\end{equation}


\begin{equation}
    R^2=\frac{\left[\sum_{i=1}^n(O_i-\mean{O}) (S_i-\mean{S})\right]^2}{\sum_{i=1}^n(O_i-\mean{O}) \sum_{i=1}^n(S_i-\mean{S})}
\end{equation}

Smaller values of RMSE and MAE and higher values of $R^2$  indicates higher model performance. 

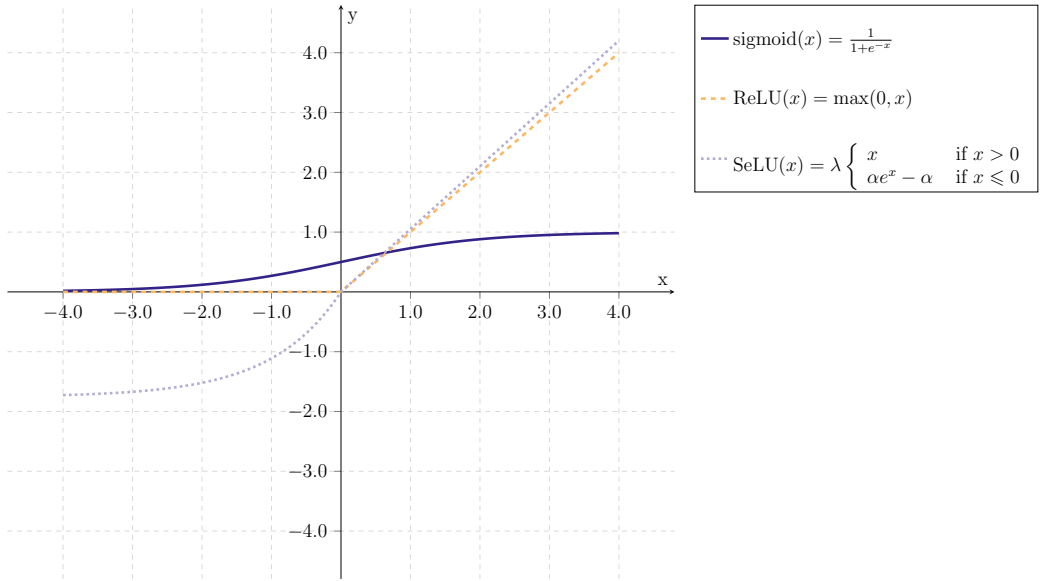
\begin{figure}[hbt!]

\begin{adjustbox}{width=1.0\textwidth,keepaspectratio}

\begin{tikzpicture}
    \definecolor{color1}{HTML}{332288}
    \definecolor{color2}{HTML}{FDB863}
    \definecolor{color3}{HTML}{B2ABD2}
    \definecolor{color4}{HTML}{5E3C99}
    \begin{axis}[
        legend style={minimum height=1.3cm},
        legend pos=outer north east,
        legend cell align={left},
        axis x line=middle,
        axis y line=middle,
        x tick label style={/pgf/number format/fixed,
                            /pgf/number format/fixed zerofill,
                            /pgf/number format/precision=1},
        y tick label style={/pgf/number format/fixed,
                            /pgf/number format/fixed zerofill,
                            /pgf/number format/precision=1},
        grid = major,
        width=16cm,
        height=14cm,
        grid style={dashed, gray!30},
        xmin=-4,     
        xmax= 4,    
        ymin=-4,     
        ymax= 4,   
        xlabel=x,
        ylabel=y,
        tick align=outside,
        enlargelimits=true]
      \addplot[domain=-4:4, color1, ultra thick,samples=500] {1/(1+exp(-x))};
      \addplot[domain=-4:4, color2, ultra thick,samples=500, dashed] {max(0, x)};
      \addplot[domain=-4:0, color3, ultra thick,samples=500, dotted] {1.05070098* (1.67326324*exp(x) - 1.67326324)};
      \addplot[domain=0:4, color3, ultra thick,samples=500, dotted] {1.05070098* x};

      \addlegendentry{$\operatorname{sigmoid}(x)=\frac{1}{1+e^{-x}}$}
      \addlegendentry{$\operatorname{ReLU}(x)=\max(0, x)$}

      \addlegendentry{$\operatorname{SeLU}(x)= \lambda \left \{
         \begin{array}{ll}{x} & {\text { if } x>0} \\ {\alpha e^{x}-\alpha} & {\text { if } x \leqslant 0}
         \end{array}
         \right.$}

    \end{axis}
\end{tikzpicture}
\end{adjustbox}
\caption{Activation Functions used in Experiments.}
\protect{\label{figure-activation-function}}

\end{figure}

\subsection{Activation functions}
\label{section-activation-functions}

As mentioned in section~\ref{section-deep-learning-methods}, differences between the DNN and ANN methods are the number of layers, activation functions, and neuron types.
Figure~\ref{figure-activation-function} shows used activation functions in present study.
Classical ANN methods mostly uses $\operatorname{sigmoid}$ function (Equation~\ref{eq-sigmoid}).

\begin{equation}
\operatorname{sigmoid}(x)=\frac{1}{1+e^{-x}}
\label{eq-sigmoid}
\end{equation}

$\operatorname{ReLU}$ function (Equation~\ref{eq-ReLU}) was made popular by DNN methods \citep{Glorot2011Deep}.
Only the previously applied DNN method \citep{Saggi2019Reference} was used $\operatorname{ReLU}$ function in the hidden layers.

\begin{equation}
\operatorname{ReLU}(x)=\max(0, x)
\label{eq-ReLU}
\end{equation}

In addition to ReLU function, SeLU function (Equation~\ref{eq-SeLU}) was also tried in the experiments. 
The SeLU function was introduced by \cite{Klambauer2017Self}.
Klambauer et al proposed the best constant values for $(\alpha,\lambda)$ as $(1.67326324,1.05070098)$, respectively.

\begin{equation}
\operatorname{SeLU}(x)= \lambda \left \{
\begin{array}{ll}{x} & {\text { if } x>0} \\ {\alpha e^{x}-\alpha} & {\text { if } x \leqslant 0}
\end{array}
\right.
\label{eq-SeLU}
\end{equation}

\subsection{Dropout layer}
\label{section-dropbox-layer}

\begin{figure}[hbt!]

 \includegraphics[width=0.9\textwidth]{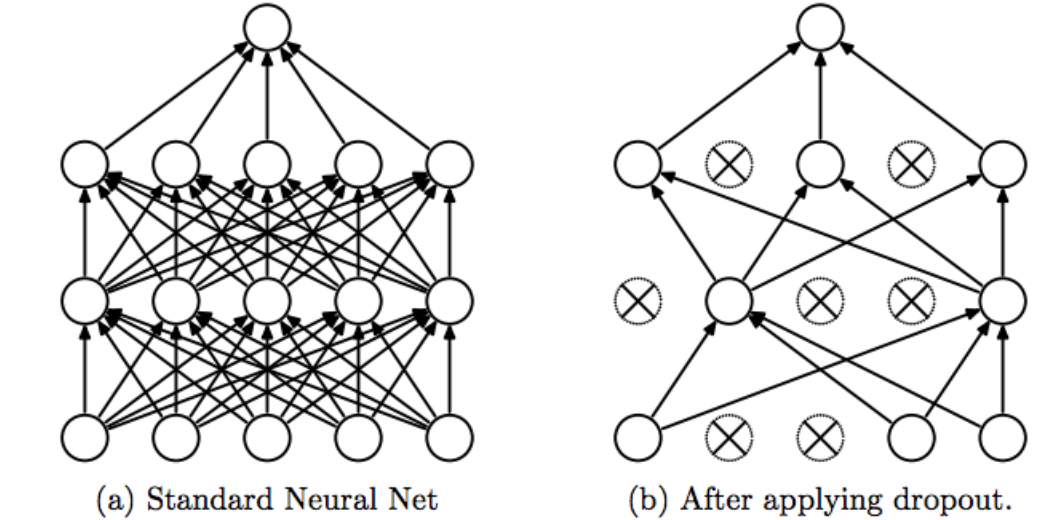}

\caption{The architecture of dropout in neural networks. \citep{Srivastava2014Dropout}}
\protect{\label{figure-dropout}}

\end{figure}

The dropout layer was first proposed by \cite{Srivastava2014Dropout}.
Dropout layer is one of the most popular solutions to over fitting problem in DNN methods.
Dropout layer is used with single parameter, dropout rate, which changes between 0 and 1.
Dropout rate controls chance of dropping out connections when neural network is training.
For example, if dropout rate is 0.5, coming neural connections to dropout layer is dropped with  50\% chance. 
An example can be seen in Figure~\ref{figure-dropout} in which connections of crossed neurons are dropped out.

To be able to see if dropout layer is helpful for over fitting  in the estimation of $ET_o$ value, different dropout rates were tried between the hidden layers.
Therefore, 3 different dropout layers were introduced in DNN models (Figure~\ref{figure-proposed-DNNs-ReLU-dropout} and Figure~\ref{figure-proposed-DNNs-SeLU-dropout}).
In the experiment, dropout rate changed from 0 to 0.5 with a rate of 0.1 increase.
In total, 6 dropout rates were tried $[0,0.1,0.2,0.3,0.4,0.5]$.
In the end, using 3 DNN models with 3 different layers and 6 dropout layers (3 DNN * 6 (Dropout Layer 2) * 6 (Dropout Layer 2) * 6 (Dropout Layer 3), 648 models were applied in the experiments.
According to our results, dropout layer is not useful for the estimation of $ET_o$ values.

\subsection{Proposed deep neural network models}
\label{section-proposed-deep-learning-model}

\begin{figure}[hbt!]
    \centering
    \begin{subfigure}{1.0\textwidth}
        \centering
        \includegraphics[width=1.0\textwidth]{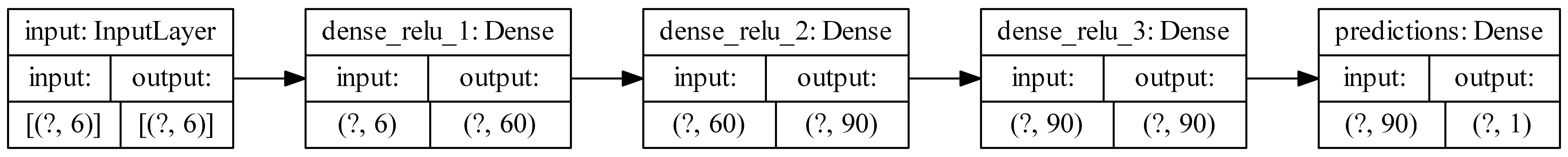}
        \caption{P-DNN-ReLU}
		\label{figure-proposed-DNNs-ReLU-nodropout}
    \end{subfigure}    
    \begin{subfigure}{1.0\textwidth}
        \centering
        \includegraphics[width=1.0\textwidth]{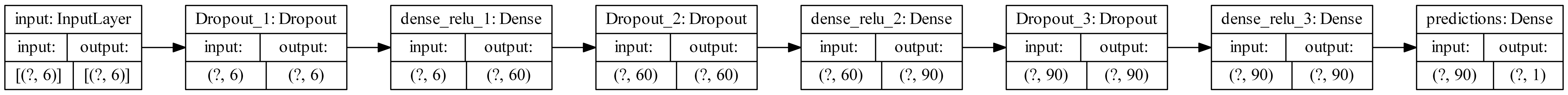}
        \caption{P-DNN-ReLU with dropout layers}
		\label{figure-proposed-DNNs-ReLU-dropout}
    \end{subfigure}
    \begin{subfigure}{1.0\textwidth}
        \centering
        \includegraphics[width=1.0\textwidth]{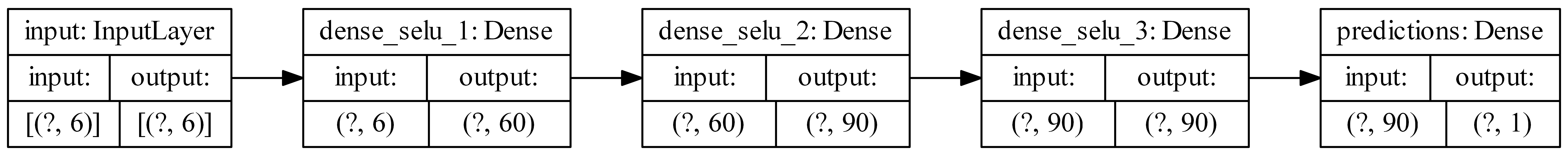}
        \caption{P-DNN-SeLU}
		\label{figure-proposed-DNNs-SeLU-nodropout}
    \end{subfigure}
    \begin{subfigure}{1.0\textwidth}
        \centering
        \includegraphics[width=1.0\textwidth]{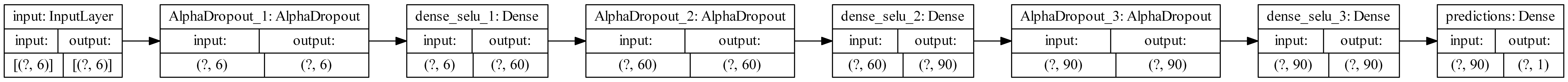}
        \caption{P-DNN-SeLU with dropout layers}
		\label{figure-proposed-DNNs-SeLU-dropout}
    \end{subfigure}
    \caption{Visualization of proposed DNN methods.}
    \label{figure-proposed-DNNs}
\end{figure}

In this study, 3 layer architecture (1-60-90-60-1) were proposed for DNN model.
For this 3 layer architecture, both ReLU and SeLU activation functions were used. Additionally, different dropout rates were also tried in the DNN model.
Figure~\ref{figure-proposed-DNNs} shows the proposed DNN models.

\section{Results and discussion}
\label{section-results}

\subsection{Comparison of all the applied models}
\label{Comparison-of-all-the-applied-models}

The best twenty performances metric of current and previous models for estimating daily $ET_o$ at the 4 meteorological stations are presented in Table~\ref{table-experimental-results-best20}. 
As can be seen in the Table~\ref{table-experimental-results-best20}, $R^2$, RMSE and MAE performance metrics ranged between 0.9913-0.9933 , 0.1811-0.2471 $mm\;day^{-1}$, 0.1333-0.1874 $mm\;day^{-1}$, respectively. 
It was found that $R^2$ values of all the applied models were higher than 0.991, indicating a strong relationship between the $ET_o$ values from the FAO-56 PM equation and those predicted by the applied models.
The RMSE and MAE values were lower than 0.25 and 0.19 $mm\;day^{-1}$, which shows excellent performance for the estimation of daily $ET_o$. 
The proposed models (P-DNN-SeLU and P-DNN-ReLU) performed better than the previous models, with $R^2$, RMSE and MAE ranging 0.9933-0.9932 , 0.1811-0.2182 $mm\;day^{-1}$, 0.1333-0.1678 $mm\;day^{-1}$, respectively.
These results confirmed that proposed DNN models are superior to previously applied ANN and DNN models for estimation of daily $ET_o$. 

\begin{table}[!htb]
\resizebox{\textwidth}{!}{%
\begin{threeparttable}[b]
\centering
\caption{The best twenty performance metric of all the currently and previously proposed models for estimation of $ET_o$ at the 4 meteorological stations according to $R^2$}
\label{table-experimental-results-best20}
\begin{tabular}{@{}llllll@{}}
\toprule
Order & model name & station name & $R^2$ & RMSE & MAE   \\ 

\midrule

01 & P-DNN-SeLU \tnote{1} & Aksaray & 0.9934 & 0.2073 & 0.1591 \\
02 & P-DNN-ReLU  & Nigde & 0.9933 & 0.2182 & 0.1678 \\
03 & P-DNN-SeLU  & Adana & 0.9932 & 0.1812 & 0.1333 \\
04 & L-DNN-Saggi  & Nigde & 0.9930 & 0.2240 & 0.1740 \\
05 & L-DNN-Saggi  & Aksaray & 0.9928 & 0.2160 & 0.1654 \\
06 & P-DNN-ReLU  & Aksaray & 0.9928 & 0.2165 & 0.1651 \\
07 & L-DNN-Saggi & Aksaray & 0.9926 & 0.2197 & 0.1683 \\
08 & L-ANN (60, 90, 60) \tnote{2} & Nigde & 0.9919 & 0.2400 & 0.1826 \\
09 & P-DNN-SeLU dropout 0 0.1 0 & Aksaray & 0.9919 & 0.2298 & 0.1788 \\
10 & L-ANN (9) & Aksaray & 0.9918 & 0.2311 & 0.1786 \\
11 & L-ANN (19) & Nigde & 0.9918 & 0.2421 & 0.1854 \\
12 & L-ANN (19) & Aksaray & 0.9916 & 0.2334 & 0.1788 \\
13 & P-DNN-SeLU dropout 0 0.1 0 & Adana & 0.9916 & 0.2023 & 0.1483 \\
14 & L-ANN (12) & Nigde & 0.9916 & 0.2453 & 0.1871 \\
15 & L-ANN (21) & Aksaray & 0.9915 & 0.2348 & 0.1796 \\
16 & L-ANN (22) & Aksaray & 0.9915 & 0.2353 & 0.1799 \\
17 & L-ANN (16) & Aksaray & 0.9915 & 0.2354 & 0.1793 \\
18 & L-ANN (28) & Aksaray & 0.9915 & 0.2355 & 0.1797 \\
19 & L-ANN (26) & Nigde & 0.9914 & 0.2472 & 0.1875 \\
20 & L-ANN (60, 90, 60) & Aksaray & 0.9913 & 0.2372 & 0.1804 \\

\bottomrule
\end{tabular}%
\begin{tablenotes}
\item [1] P- means currently proposed, L mean Literature.
\item [2] ANN(..) refers to Artificial Neural Network hidden layer neuron counts, for example (21) means 1 hidden layer with 21 neurons while (60,90,60) means 3 hidden layers with 60, 90 and 60 neurons.
\end{tablenotes}
\end{threeparttable}
}
\end{table}

However, the performance of the proposed activation SeLU decreased when dropout layer used (proposed activation SeLU dropout 0 0.1 0). 
This means that using dropout layers did not improve the modeling performances for the estimation of daily $ET_o$. 

In general, it is observed that among the previously applied models, the Saggi and Jain model had the highest performance based on the performance metrics. 
Among the stations in the top twenty best performing models, Aksaray and Niğde stations located in the semi-arid region were the most appearing, while Adana and Isparta stations located in Mediterranean region were the least appearing. 
This can be explained by the fact that the models performed better in semi arid region than the Mediterranean region.

\subsection{Comparison of proposed and previously applied ANN and DNN models}
\label{Comparison-of-proposed-and-previously-applied-ANN-and-DNN-models}

\begin{table}[!htb]
\caption{The highest performances of previously proposed methods for $ET_o$ estimation according to $R^2$ score}
\label{table-experimental-results-previous}

\begin{adjustbox}{max width=\textwidth,max totalheight=\textheight,keepaspectratio}

\begin{tabular}{p{0.02\textwidth}p{0.3\textwidth}p{0.1\textwidth}p{0.1\textwidth}p{0.1\textwidth}p{0.1\textwidth}p{0.1\textwidth}}

\toprule
& Study & ANN Hidden Layers & Best ($R^2$) Adana & Best ($R^2$) Aksaray & Best ($R^2$) Isparta & Best ($R^2$) Nigde\\ 

\midrule

1 & \cite{Landeras2008Comparison},\cite{Traore2010Artificial},\cite{Rahimikhoob2014Comparison},\cite{Shiri2014Comparison},\cite{Yassin2016Artificial},\cite{Kisi2016investigation},\cite{Antonopoulos2017Daily},\cite{Dou2018Evapotranspiration} & (1-30) & 0.9864  & 0.9918 & 0.9880 &  0.9918 \\

2 & \cite{Huo2012Artificial} &  4-5, 5-6 & 0.9814 & 0.9883 & 0.9837 &  0.9885 \\

3 & \cite{Saggi2019Reference} & 40-60-40 & 0.9905  & 0.9928 & 0.9896 &  0.9931 \\

3 & Proposed ReLU & 60-90-60 & 0.9888  & 0.9928 & 0.9890 &  0.9934 \\

4 & Proposed SeLU & 60-90-60 & 0.9925  & 0.9940 & 0.9887 &  0.9936 \\
 \bottomrule

\end{tabular}
\end{adjustbox}
\end{table}

The comparison results of the previous models (ANN and DNN) and two current DNN models can be seen in Table~\ref{table-experimental-results-previous}.
The table showed only the best performances among the experiments.
According to Table~\ref{table-experimental-results-previous}, it was seen that $R^2$ values of all the previously applied and proposed models were higher than 0.98.
DNN models compared to ANN models improved $R^2$ values in the 3-4 decimal points.
If the complexity of the models and time to train are taken into account, these results may be considered as diminishing improvements according to requirements for $ET_o$.

\subsection{Comparison of meteorological station performances}
\label{Comparison-of-meteorological-station-performances}

\begin{table}[hbt!]
\centering
\caption{The best five performance metric of proposed and previously applied models for estimation of $ET_o$ at the Adana, Aksaray, Isparta and Niğde stations respectively according to $R^2$ score}
\label{table-best-5-results-stations}
\resizebox{\textwidth}{!}{%
\begin{tabular}{@{}llllll@{}}
\toprule
Rank & model name & station name & $R^2$ & RMSE & MAE  \\
 \midrule

01 & P-DNN-SeLU  & Adana & 0.9925 & 0.1912 & 0.1409 \\
02 & P-DNN-SeLU dropout 0.0 0.1 0.0 & Adana & 0.9916 & 0.2023 & 0.1483 \\
03 & L-DNN-Saggi  & Adana & 0.9905 & 0.2151 & 0.1520 \\
04 & P-DNN-ReLU  & Adana & 0.9888 & 0.2336 & 0.1664 \\
05 & ANN (60, 90, 60) & Adana & 0.9865 & 0.2557 	 & 0.1860 \\

 \midrule

01 & P-DNN-SeLU  & Aksaray & 0.9940 & 0.1977 & 0.1503 \\
02 & L-DNN-Saggi  & Aksaray & 0.9928 & 0.2160 & 0.1654 \\
03 & P-DNN-ReLU  & Aksaray & 0.9928 & 0.2165 & 0.1652 \\
04 & P-DNN-SeLU dropout 0.0 0.1 0.0 & Aksaray & 0.9919 & 0.2298 & 0.1788 \\
05 & ANN (9) & Aksaray & 0.9918 & 0.2311 & 0.1786 \\
 \midrule

01 & L-DNN-Saggi  & Isparta & 0.9896 & 0.2346 & 0.1779 \\
02 & P-DNN-ReLU  & Isparta & 0.9890 & 0.2405 & 0.1848 \\
03 & P-DNN-SeLU  & Isparta & 0.9887 & 0.2438 & 0.1864 \\
04 & ANN (16) & Isparta & 0.9880 & 0.2513 & 0.1942 \\
05 & P-DNN-SeLU dropout 0.0 0.0 0.1 & Isparta & 0.9878 & 0.2538 & 0.1938 \\

 \midrule
01 & P-DNN-SeLU  & Nigde & 0.9936 & 0.2132 & 0.1644 \\
02 & P-DNN-ReLU  & Nigde & 0.9934 & 0.2168 & 0.1681 \\
03 & L-DNN-Saggi  & Nigde & 0.9931 & 0.2219 & 0.1708 \\
04 & ANN (60, 90, 60) & Nigde & 0.9919 & 0.2400 & 0.1826 \\
05 & ANN (19) & Nigde & 0.9918 & 0.2421 & 0.1854 \\
\bottomrule

\end{tabular}%
}
\end{table}

The best five performance metric of currently proposed and previous models for estimation of $ET_o$ at the Adana, Aksaray, Isparta and Niğde stations are presented in Table~\ref{table-best-5-results-stations}. 
In general, Table~\ref{table-best-5-results-stations} showed that the P-DNN-SeLU model had the highest performance in Aksaray, Niğde and Adana stations.
In that case, $R^2$ values of P-DNN-SeLU model are 0.9939, 0.9936 and 0.9924, RMSE values are 0.1977, 0.2131 and 0.1911, MAE values are 0.1502, 0.1643 and 0.1409. 
However, Saggi and Jain model had the highest performance in Isparta station with the value of $R^2$ to 0.9896, RMSE to 0.2346 and MAE to 0.1779, respectively.
Obviously, the proposed activation SeLU model in Aksaray and Niğde stations had slightly better prediction accuracy than the P-DNN-SeLU model in Adana and Isparta stations. 
It can be seen that the P-DNN-SeLU model showed more improvements in daily $ET_o$ estimation in the semi arid region, compared with those in the Mediterranean regions.

\begin{figure}[hbt!]

\includegraphics[width=1.2\textwidth]{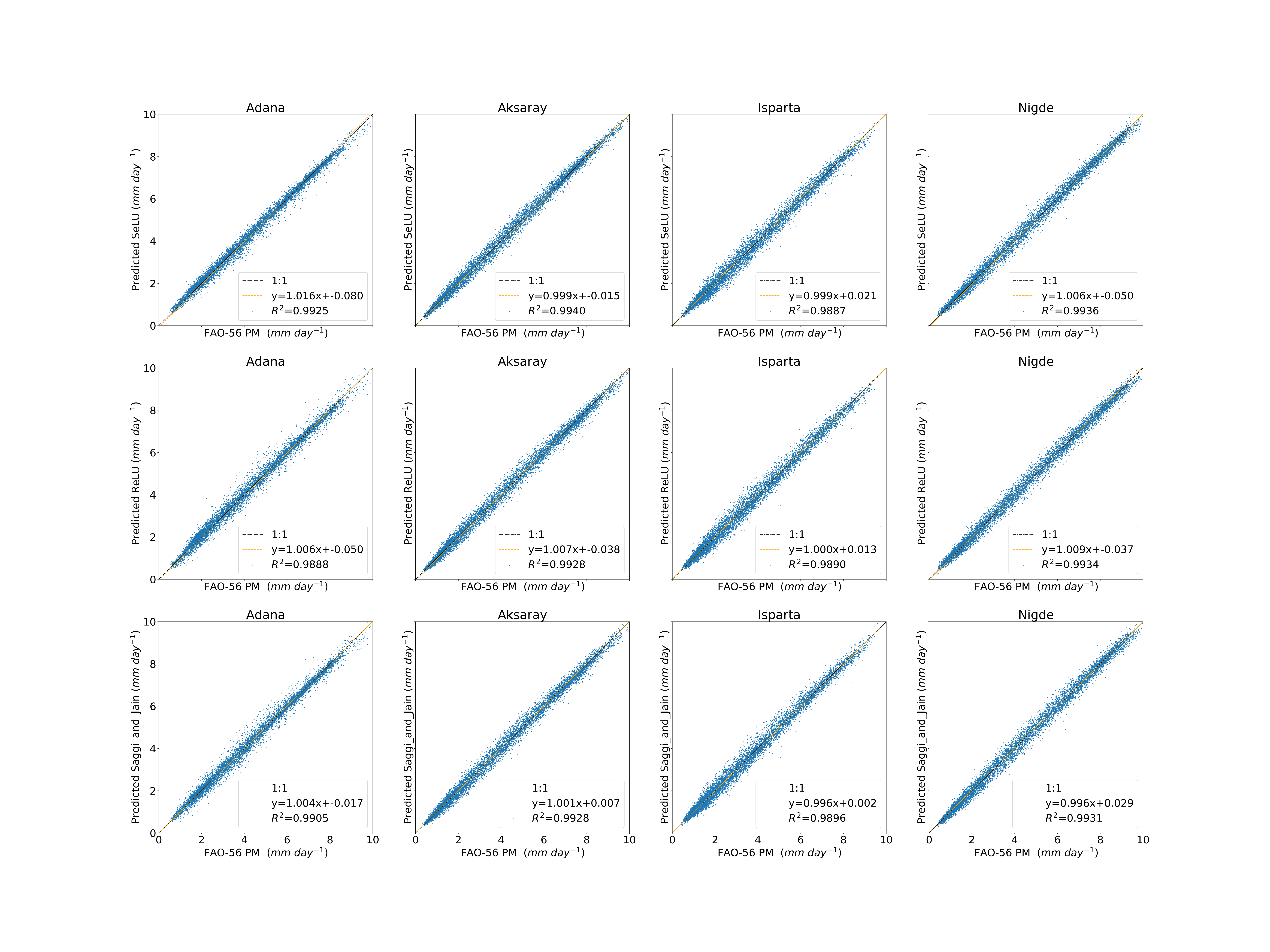}

\caption{Scatter plots for 3 Deep Neural Networks.}
\label{figure-scatter-plot}

\end{figure}

The scatter plot of estimated $ET_o$ values by P-DNN-SeLU, P-DNN-ReLU and Saggi and Jain models compared with the FAO-56 PM values at the Adana, Aksaray, Isparta and Niğde stations are presented in Figure~\ref{figure-scatter-plot}. 
The figure showed that the plotted data points mostly correlated close towards the 1:1 line. 
However, the models in Isparta station yielded more scattered daily $ET_o$ values compared to other three stations. 
The models were more close to those obtained with FAO-56 PM equation in Aksaray station. 
These results indicated that the models showed a much higher prediction accuracy of daily $ET_o$ value in Aksaray station.

\section{Conclusions}
\label{section-conclusions}

This study assessed the application of 14 ANN methods (2 new DNN, 1 previous DNN and 11 previous ANN) for estimation of daily $ET_o$ in the two different climate zones of Turkey. 
The models used 6 input meteorological data including $T_{max}$, $T_{min}$, $R_n$, $RH_{max}$, $RH_{min}$ and $U_2$ from 4 weather stations (Adana, Aksaray, Isparta and Niğde) during 1999-2018 in Turkey. 
The results demonstrated that all models of DNN and ANN achieved satisfactory accuracy for estimation of daily $ET_o$ using available meteorological data.
Especially, the DNN methods were highly effective in estimating  $ET_o$ value.
It can be seen from the results that the performance of the models became more reliable when cross validation was used in the study. 
However, the result showed that dropout layer was not useful for $ET_o$ estimation. 
In addition, using more powerful architecture did not improve the estimation of $ET_o$.

In general, among the meteorological stations, Aksaray station offered the best prediction accuracy, while Isparta station performed the least prediction accuracy in all the DNN and ANN models.
The overall results indicated that the proposed model of P-DNN-SeLU made a significant improvement in accuracy among the other models.
Therefore, P-DNN-SeLU model has a very high potential for estimation of daily $ET_o$ in different climatic zones of Turkey, even possibly in other zones of the world.
In short, P-DNN-SeLU model could be applied in the future studies to estimate $ET_o$ under different climate conditions. 
Finally, further studies should be carried out to evaluate the applicability of P-DNN-SeLU model under limited input data, in places where meteorological variables are limited.




\todo[inline]{scatter plot - hepsi için, supplementary material 5xx scatter plot bir dosyada.}

\todo[inline]{add Critical Difference Diagrams: https://github.com/hfawaz/cd-diagram}

\todo[inline]{Figure 2: Visualization of the artificial neuron. Kalite artırma}

\todo[inline]{arxiv upload}

\todo[inline]{Journal of Hydrology submit}


\bibliographystyle{elsarticle-harv}

\begin{thebibliography}{53}
\expandafter\ifx\csname natexlab\endcsname\relax\def\natexlab#1{#1}\fi
\expandafter\ifx\csname url\endcsname\relax
  \def\url#1{\texttt{#1}}\fi
\expandafter\ifx\csname urlprefix\endcsname\relax\def\urlprefix{URL }\fi

\bibitem[{Ahooghalandari et~al.(2016)Ahooghalandari, Khiadani, and
  Jahromi}]{Ahooghalandari2016Developing}
Ahooghalandari, M., Khiadani, M., Jahromi, M.~E., 2016. Developing equations
  for estimating reference evapotranspiration in australia. Water Resources
  Management.

\bibitem[{Allen et~al.(1998)Allen, Pereira, Raes, and Smith}]{Allen1998Crop}
Allen, R., Pereira, L., Raes, D., Smith, M., 1998. Crop evapotranspiration
  guidelines for computing crop water requirements-fao irrigation and drainage
  paper 56. Tech. rep., Food and Agriculture Organization (FAO).

\bibitem[{Amodei et~al.(2016)}]{Amodei2016Deep}
Amodei, D., et~al., 2016. Deep speech 2 : End-to-end speech recognition in
  english and mandarin. In: Proceedings of The 33rd International Conference on
  Machine Learning.

\bibitem[{Antonopoulos and Antonopoulos(2017)}]{Antonopoulos2017Daily}
Antonopoulos, V.~Z., Antonopoulos, A.~V., 2017. Daily reference
  evapotranspiration estimates by artificial neural networks technique and
  empirical equations using limited input climate variables. Computers and
  Electronics in Agriculture 132, 86 -- 96.
\newline\urlprefix\url{http://www.sciencedirect.com/science/article/pii/S0168169916305506}

\bibitem[{Bui et~al.(2020)Bui, Nguyen, Nguyen, Pham, Nguyen, and
  Pham}]{Bui2020Verification}
Bui, Q.-T., Nguyen, Q.-H., Nguyen, X.~L., Pham, V.~D., Nguyen, H.~D., Pham,
  V.-M., 2020. Verification of novel integrations of swarm intelligence
  algorithms into deep learning neural network for flood susceptibility
  mapping. Journal of Hydrology 581, 124379.
\newline\urlprefix\url{http://www.sciencedirect.com/science/article/pii/S002216941931114X}

\bibitem[{Dinpashoh(2006)}]{Dinpashoh2006Study}
Dinpashoh, Y., 2006. Study of reference crop evapotranspiration in i.r. of
  iran. Agricultural Water Management.

\bibitem[{Dou and Yang(2018)}]{Dou2018Evapotranspiration}
Dou, X., Yang, Y., 2018. Evapotranspiration estimation using four different
  machine learning approaches in different terrestrial ecosystems. Computers
  and Electronics in Agriculture 148, 95 -- 106.
\newline\urlprefix\url{http://www.sciencedirect.com/science/article/pii/S016816991731476X}

\bibitem[{Dyrmann et~al.(2016)Dyrmann, Karstoft, and
  Midtiby}]{Dyrmann2016Plant}
Dyrmann, M., Karstoft, H., Midtiby, H.~S., 2016. Plant species classification
  using deep convolutional neural network. Biosystems Engineering 151, 72 --
  80.
\newline\urlprefix\url{http://www.sciencedirect.com/science/article/pii/S1537511016301465}

\bibitem[{Fan et~al.(2019)Fan, Ma, Wu, Zhang, Yu, and Zeng}]{Fan2019Light}
Fan, J., Ma, X., Wu, L., Zhang, F., Yu, X., Zeng, W., 2019. Light gradient
  boosting machine: An efficient soft computing model for estimating daily
  reference evapotranspiration with local and external meteorological data.
  Agricultural Water Management 225, 105758.

\bibitem[{Fan et~al.(2018)Fan, Yue, Wu, Zhang, Cai, Wang, Lu, and
  Xiang}]{Fan2018Evaluation}
Fan, J., Yue, W., Wu, L., Zhang, F., Cai, H., Wang, X., Lu, X., Xiang, Y.,
  2018. Evaluation of svm, elm and four tree-based ensemble models for
  predicting daily reference evapotranspiration using limited meteorological
  data in different climates of china. Agricultural and Forest Meteorology 263,
  225 -- 241.
\newline\urlprefix\url{http://www.sciencedirect.com/science/article/pii/S0168192318302855}

\bibitem[{Feng et~al.(2017)Feng, Cui, Gong, Zhang, and
  Zhao}]{Feng2017Evaluation}
Feng, Y., Cui, N., Gong, D., Zhang, Q., Zhao, L., 2017. Evaluation of random
  forests and generalized regression neural networks for daily reference
  evapotranspiration modelling. Agricultural Water Management 193, 163 -- 173.
\newline\urlprefix\url{http://www.sciencedirect.com/science/article/pii/S0378377417302597}

\bibitem[{Feng et~al.(2016)Feng, Cui, Zhao, Hu, and Gong}]{Feng2016Comparison}
Feng, Y., Cui, N., Zhao, L., Hu, X., Gong, D., 2016. Comparison of elm, gann,
  wnn and empirical models for estimating reference evapotranspiration in humid
  region of southwest china. Journal of Hydrology 536, 376 -- 383.
\newline\urlprefix\url{http://www.sciencedirect.com/science/article/pii/S0022169416300981}

\bibitem[{Ferreira and da~Cunha(2020)}]{Ferreira2020New}
Ferreira, L.~B., da~Cunha, F.~F., may 2020. New approach to estimate daily
  reference evapotranspiration based on hourly temperature and relative
  humidity using machine learning and deep learning. Agricultural Water
  Management 234, 106113.

\bibitem[{Ferreira et~al.(2019)Ferreira, da~Cunha, de~Oliveira, and
  Filho}]{Ferreira2019Estimation}
Ferreira, L.~B., da~Cunha, F.~F., de~Oliveira, R.~A., Filho, E. I.~F., 2019.
  Estimation of reference evapotranspiration in brazil with limited
  meteorological data using ann and svm – a new approach. Journal of
  Hydrology 572, 556 -- 570.
\newline\urlprefix\url{http://www.sciencedirect.com/science/article/pii/S0022169419302689}

\bibitem[{Glorot et~al.(2011)Glorot, Bordes, and Bengio}]{Glorot2011Deep}
Glorot, X., Bordes, A., Bengio, Y., 2011. Deep sparse rectifier neural
  networks. In: AISTATS.

\bibitem[{Gocić et~al.(2015)Gocić, Motamedi, Shamshirband, Petković, Ch,
  Hashim, and Arif}]{Gocic2015Soft}
Gocić, M., Motamedi, S., Shamshirband, S., Petković, D., Ch, S., Hashim, R.,
  Arif, M., 2015. Soft computing approaches for forecasting reference
  evapotranspiration. Computers and Electronics in Agriculture 113, 164 -- 173.
\newline\urlprefix\url{http://www.sciencedirect.com/science/article/pii/S0168169915000526}

\bibitem[{Golhani et~al.(2018)Golhani, Balasundram, Vadamalai, and
  Pradhan}]{Golhani2018review}
Golhani, K., Balasundram, S.~K., Vadamalai, G., Pradhan, B., 2018. A review of
  neural networks in plant disease detection using hyperspectral data.
  Information Processing in Agriculture 5~(3), 354 -- 371.
\newline\urlprefix\url{http://www.sciencedirect.com/science/article/pii/S2214317317301774}

\bibitem[{Grinblat et~al.(2016)Grinblat, Uzal, Larese, and
  Granitto}]{Grinblat2016Deep}
Grinblat, G.~L., Uzal, L.~C., Larese, M.~G., Granitto, P.~M., 2016. Deep
  learning for plant identification using vein morphological patterns.
  Computers and Electronics in Agriculture 127, 418 -- 424.
\newline\urlprefix\url{http://www.sciencedirect.com/science/article/pii/S0168169916304665}

\bibitem[{Guo et~al.(2014)Guo, Singh, Lee, Lewis, and Wang}]{Guo2014Deep}
Guo, X., Singh, S., Lee, H., Lewis, R.~L., Wang, X., 2014. Deep learning for
  real-time atari game play using offline monte-carlo tree search planning. In:
  Advances in Neural Information Processing Systems 27. Curran Associates, Inc.
\newline\urlprefix\url{http://papers.nips.cc/paper/5421-deep-learning-for-real-time-atari-game-play-using-offline-monte-carlo-tree-search-planning.pdf}

\bibitem[{Hall et~al.(2009)Hall, Frank, Holmes, Pfahringer, Reutemann, and
  Witten}]{Hall2009WEKA}
Hall, M., Frank, E., Holmes, G., Pfahringer, B., Reutemann, P., Witten, I.~H.,
  2009. The weka data mining software: An update. SIGKDD Explor. Newsl.

\bibitem[{Hargreaves and Samani(1985)}]{Hargreaves1985Reference}
Hargreaves, G., Samani, Z., 1985. Reference crop evapotranspiration from
  temperature. Applied Engineering in Agriculture.

\bibitem[{Huang et~al.(2019)Huang, Wu, Ma, Zhang, Fan, Yu, Zeng, and
  Zhou}]{Huang2019Evaluation}
Huang, G., Wu, L., Ma, X., Zhang, W., Fan, J., Yu, X., Zeng, W., Zhou, H.,
  2019. Evaluation of catboost method for prediction of reference
  evapotranspiration in humid regions. Journal of Hydrology 574, 1029 -- 1041.
\newline\urlprefix\url{http://www.sciencedirect.com/science/article/pii/S0022169419304251}

\bibitem[{Huo et~al.(2012)Huo, Feng, Kang, and Dai}]{Huo2012Artificial}
Huo, Z., Feng, S., Kang, S., Dai, X., 2012. Artificial neural network models
  for reference evapotranspiration in an arid area of northwest china. Journal
  of Arid Environments 82, 81 -- 90.

\bibitem[{Irmak et~al.(2003)Irmak, Irmak, Allen, and Jones}]{Irmak2003Solar}
Irmak, S., Irmak, A., Allen, R.~G., Jones, J.~W., 2003. Solar and net
  radiation-based equations to estimate reference evapotranspiration in humid
  climates. Journal of Irrigation and Drainage Engineering.

\bibitem[{Kamilaris and Prenafeta-Boldú(2018)}]{Kamilaris2018Deep}
Kamilaris, A., Prenafeta-Boldú, F.~X., 2018. Deep learning in agriculture: A
  survey. Computers and Electronics in Agriculture 147, 70 -- 90.
\newline\urlprefix\url{http://www.sciencedirect.com/science/article/pii/S0168169917308803}

\bibitem[{Kisi(2015)}]{Kisi2015Pan}
Kisi, O., 2015. Pan evaporation modeling using least square support vector
  machine, multivariate adaptive regression splines and m5 model tree. Journal
  of Hydrology.

\bibitem[{Kisi and Kilic(2016)}]{Kisi2016investigation}
Kisi, O., Kilic, Y., 2016. An investigation on generalization ability of
  artificial neural networks and m5 model tree in modeling reference
  evapotranspiration. Theoretical and Applied Climatology 126~(3), 413--425.

\bibitem[{{Klambauer} et~al.(2017){Klambauer}, {Unterthiner}, {Mayr}, and
  {Hochreiter}}]{Klambauer2017Self}
{Klambauer}, G., {Unterthiner}, T., {Mayr}, A., {Hochreiter}, S., 2017.
  Self-normalizing neural networks. CoRR.

\bibitem[{Kottek et~al.(2006)Kottek, Grieser, Beck, Rudolf, and
  Rubel}]{Kottek2006World}
Kottek, M., Grieser, J., Beck, C., Rudolf, B., Rubel, F., 2006. World map of
  the köppen-geiger climate classification updated. Meteorologische
  Zeitschrift 15, 259--263.

\bibitem[{Landeras et~al.(2008)Landeras, Ortiz-Barredo, and
  López}]{Landeras2008Comparison}
Landeras, G., Ortiz-Barredo, A., López, J.~J., 2008. Comparison of artificial
  neural network models and empirical and semi-empirical equations for daily
  reference evapotranspiration estimation in the basque country (northern
  spain). Agricultural Water Management.

\bibitem[{LeCun et~al.(2015)LeCun, Bengio, and Hinton}]{LeCun2015Deep}
LeCun, Y., Bengio, Y., Hinton, G., May 2015. Deep learning. Nature 521, 436.
\newline\urlprefix\url{https://doi.org/10.1038/nature14539}

\bibitem[{Lee et~al.(2020)Lee, Shin, Kim, and Singh}]{Lee2020Stochastic}
Lee, T., Shin, J.-Y., Kim, J.-S., Singh, V.~P., 2020. Stochastic simulation on
  reproducing long-term memory of hydroclimatological variables using deep
  learning model. Journal of Hydrology 582, 124540.
\newline\urlprefix\url{http://www.sciencedirect.com/science/article/pii/S0022169419312752}

\bibitem[{López-Urrea et~al.(2006)López-Urrea, de~Santa~Olalla, Fabeiro, and
  Moratalla}]{Lopez-Urrea2006Testing}
López-Urrea, R., de~Santa~Olalla, F.~M., Fabeiro, C., Moratalla, A., 2006.
  Testing evapotranspiration equations using lysimeter observations in a
  semiarid climate. Agricultural Water Management.

\bibitem[{Nielsen(2015)}]{Nielsen2015Neural}
Nielsen, M., 2015. Neural Networks and Deep Learning. Determination press.
\newline\urlprefix\url{http://neuralnetworksanddeeplearning.com/index.html}

\bibitem[{Nwankpa et~al.(2018)Nwankpa, Ijomah, Gachagan, and
  Marshall}]{Nwankpa2018Activation}
Nwankpa, C., Ijomah, W., Gachagan, A., Marshall, S., 2018. Activation
  functions: Comparison of trends in practice and research for deep learning.
  CoRR.

\bibitem[{Peng et~al.(2017)Peng, Li, and Feng}]{Peng2017best}
Peng, L., Li, Y., Feng, H., 12 2017. The best alternative for estimating
  reference crop evapotranspiration in different sub-regions of mainland china.
  Scientific Reports 7.

\bibitem[{Pereira et~al.(2015)Pereira, Allen, Smith, and
  Raes}]{Pereira2015Crop}
Pereira, L.~S., Allen, R.~G., Smith, M., Raes, D., 2015. Crop
  evapotranspiration estimation with fao56: Past and future. Agricultural Water
  Management.

\bibitem[{Priestley and Taylor(1972)}]{Priestley1972Assessment}
Priestley, C. H.~B., Taylor, R.~J., 1972. On the assessment of surface heat
  flux and evaporation using large-scale parameters. Monthly Weather Review
  100~(2), 81--92.

\bibitem[{Rahimikhoob(2014)}]{Rahimikhoob2014Comparison}
Rahimikhoob, A., 02 2014. Comparison between m5 model tree and neural networks
  for estimating reference evapotranspiration in an arid environment. Water
  Resources Management 28, 657--669.

\bibitem[{Saggi and Jain(2019)}]{Saggi2019Reference}
Saggi, M.~K., Jain, S., 2019. Reference evapotranspiration estimation and
  modeling of the punjab northern india using deep learning. Computers and
  Electronics in Agriculture 156, 387 -- 398.
\newline\urlprefix\url{http://www.sciencedirect.com/science/article/pii/S0168169918308779}

\bibitem[{Shiri et~al.(2014)Shiri, Nazemi, Sadraddini, Landeras, Kisi, Fard,
  and Marti}]{Shiri2014Comparison}
Shiri, J., Nazemi, A.~H., Sadraddini, A.~A., Landeras, G., Kisi, O., Fard,
  A.~F., Marti, P., 2014. Comparison of heuristic and empirical approaches for
  estimating reference evapotranspiration from limited inputs in iran.
  Computers and Electronics in Agriculture.

\bibitem[{Srivastava et~al.(2014)Srivastava, Hinton, Krizhevsky, Sutskever, and
  Salakhutdinov}]{Srivastava2014Dropout}
Srivastava, N., Hinton, G., Krizhevsky, A., Sutskever, I., Salakhutdinov, R.,
  2014. Dropout: A simple way to prevent neural networks from overfitting.
  Journal of Machine Learning Research 15, 1929--1958.
\newline\urlprefix\url{http://jmlr.org/papers/v15/srivastava14a.html}

\bibitem[{Tao et~al.(2018)Tao, Diop, Bodian, Djaman, Ndiaye, and
  Yaseen}]{Tao2018Reference}
Tao, H., Diop, L., Bodian, A., Djaman, K., Ndiaye, P.~M., Yaseen, Z.~M., 2018.
  Reference evapotranspiration prediction using hybridized fuzzy model with
  firefly algorithm: Regional case study in burkina faso. Agricultural Water
  Management 208, 140 -- 151.
\newline\urlprefix\url{http://www.sciencedirect.com/science/article/pii/S0378377418308151}

\bibitem[{Torres et~al.(2011)Torres, Walker, and McKee}]{Torres2011Forecasting}
Torres, A.~F., Walker, W.~R., McKee, M., 2011. Forecasting daily potential
  evapotranspiration using machine learning and limited climatic data.
  Agricultural Water Management.

\bibitem[{Trabert(2019)}]{Trabert2019Neue}
Trabert, W., 09 2019. Neue beobachtungen über verdampfungsgeschwindigkeiten.
  Meteorol. Z. 13, 261--263.

\bibitem[{Traore et~al.(2010)Traore, Wang, and Kerh}]{Traore2010Artificial}
Traore, S., Wang, Y.-M., Kerh, T., 2010. Artificial neural network for modeling
  reference evapotranspiration complex process in sudano-sahelian zone.
  Agricultural Water Management 97~(5), 707 -- 714.

\bibitem[{Valiantzas(2013)}]{Valiantzas2013Simplified}
Valiantzas, J.~D., 2013. Simplified forms for the standardized fao-56
  penman–monteith reference evapotranspiration using limited weather data.
  Journal of Hydrology.

\bibitem[{Wang et~al.(2020)Wang, Zhang, Chang, and Li}]{Wang2020Deep}
Wang, N., Zhang, D., Chang, H., Li, H., 2020. Deep learning of subsurface flow
  via theory-guided neural network. Journal of Hydrology 584, 124700.
\newline\urlprefix\url{http://www.sciencedirect.com/science/article/pii/S0022169420301608}

\bibitem[{Wu et~al.(2019)Wu, Zhou, Ma, Fan, and Zhang}]{Wu2019Daily}
Wu, L., Zhou, H., Ma, X., Fan, J., Zhang, F., 2019. Daily reference
  evapotranspiration prediction based on hybridized extreme learning machine
  model with bio-inspired optimization algorithms: Application in contrasting
  climates of china. Journal of Hydrology, 123960.
\newline\urlprefix\url{http://www.sciencedirect.com/science/article/pii/S0022169419306808}

\bibitem[{Yamaç et~al.(2020)Yamaç, Şeker, and Negiş}]{Yamac2020Evaluation}
Yamaç, S.~S., Şeker, C., Negiş, H., 2020. Evaluation of machine learning
  methods to predict soil moisture constants with different combinations of
  soil input data for calcareous soils in a semi arid area. Agricultural Water
  Management 234, 106121.
\newline\urlprefix\url{http://www.sciencedirect.com/science/article/pii/S037837741932356X}

\bibitem[{Yamaç and Todorovic(2020)}]{Yamac2020Estimation}
Yamaç, S.~S., Todorovic, M., 2020. Estimation of daily potato crop
  evapotranspiration using three different machine learning algorithms and four
  scenarios of available meteorological data. Agricultural Water Management.

\bibitem[{Yassin et~al.(2016)Yassin, Alazba, and Mattar}]{Yassin2016Artificial}
Yassin, M.~A., Alazba, A., Mattar, M.~A., 2016. Artificial neural networks
  versus gene expression programming for estimating reference
  evapotranspiration in arid climate. Agricultural Water Management.

\bibitem[{{Young} et~al.(2018){Young}, {Hazarika}, {Poria}, and
  {Cambria}}]{Young2018Recent}
{Young}, T., {Hazarika}, D., {Poria}, S., {Cambria}, E., 2018. Recent trends in
  deep learning based natural language processing [review article]. IEEE
  Computational Intelligence Magazine.

\end{thebibliography}

\end{document}